\documentclass[10pt,twocolumn,letterpaper]{article}

\usepackage{iccv}              
\usepackage{multirow}
\usepackage{multibib}
\usepackage[separate-uncertainty = true]{siunitx}

\definecolor{iccvblue}{rgb}{0.21,0.49,0.74}
\usepackage[pagebackref,breaklinks,colorlinks,allcolors=iccvblue]{hyperref}

\def\paperID{11725} 
\def\confName{ICCV}
\def\confYear{2025}

\usepackage{soul}
\usepackage{adjustbox}
\usepackage{scalerel,graphicx,xparse}
\usepackage{pifont}
\usepackage{tabularx}
\usepackage{multirow}
\usepackage{amsmath}
\usepackage{bm}
\usepackage{amsfonts} 
\usepackage{amssymb}
\usepackage{xcolor,colortbl}
\usepackage[separate-uncertainty = true]{siunitx}

\NewDocumentCommand\purpleumbrella{}{\scalerel*{\includegraphics{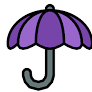}}{X}}
\NewDocumentCommand\blueumbrella{}{\scalerel*{\includegraphics{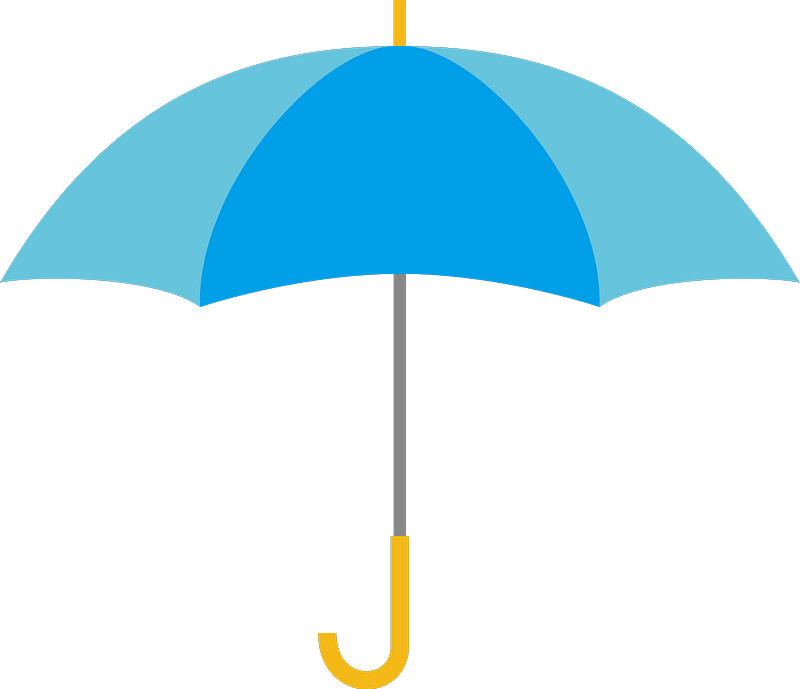}}{X}}
\NewDocumentCommand\bluebox{}{\scalerel*{\includegraphics{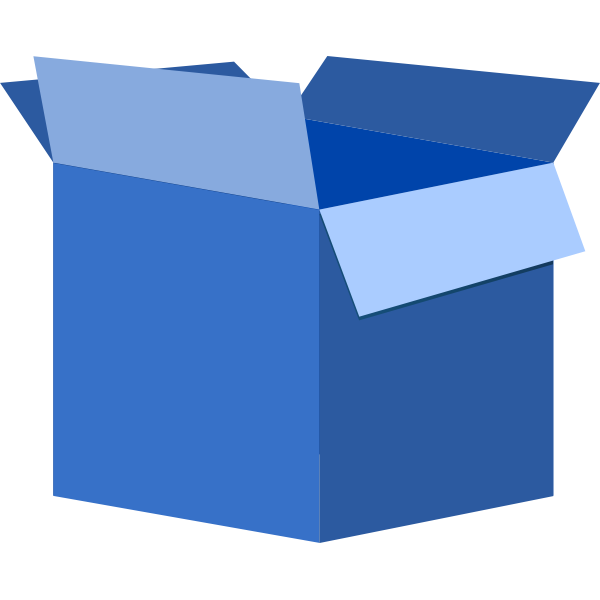}}{X}}
\NewDocumentCommand\greenbox{}{\scalerel*{\includegraphics{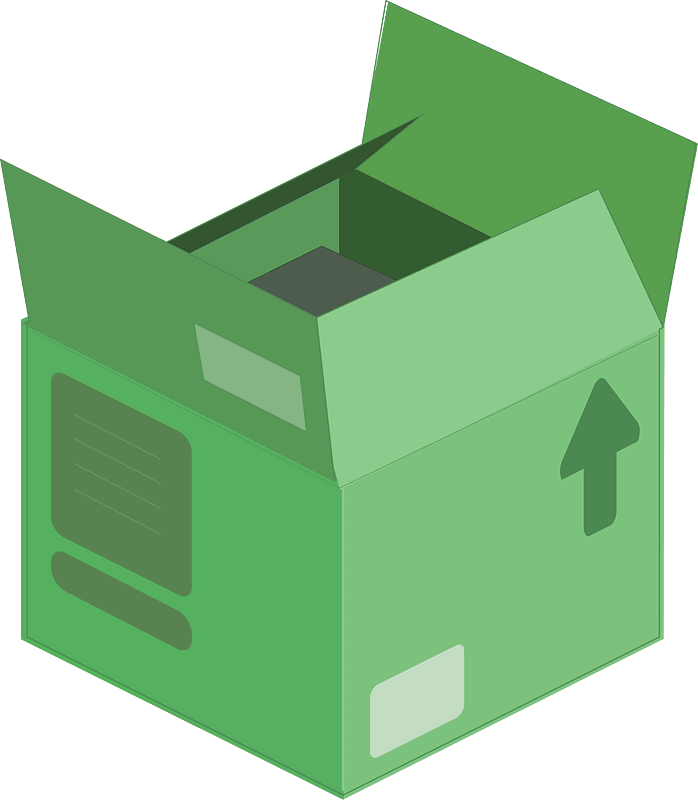}}{X}}
\NewDocumentCommand\trafficcone{}{\scalerel*{\includegraphics{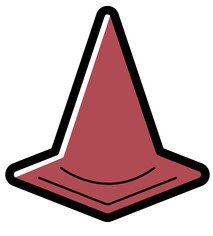}}{X}}

\newcommand{\bad}[1]{\textcolor{red!75!black}{#1}}
\newcommand{\pluseq}{\mathrel{+}=}

\newcommand{\vgPtoW}{$\textit{SegVisualGuide}$~($\textit{path}_{P_{t}\rightarrow \omega^*}$)}

\newcommand*\samethanks[1][\value{footnote}]{\footnotemark[#1]}
\makeatletter
\def\@fnsymbol#1{\ensuremath{\ifcase#1\or ^1\or ^2\or ^*\else\@ctrerr\fi}}
\makeatother
\title{Virtual Guidance as a Mid-level Representation for Navigation\\with Augmented Reality}

\author{Hsuan-Kung Yang\thanks{National Tsing Hua University} \and 
Tsung-Chih Chiang\textsuperscript{*}\samethanks[1] \and 
Jou-Min Liu\textsuperscript{*}\samethanks[1] \and 
Ting-Ru Liu\textsuperscript{*}\samethanks[1] \and
Chun-Wei Huang\textsuperscript{*}\samethanks[1] \and 
Tsu-Ching Hsiao\samethanks[1] \and
Chun-Yi Lee\thanks{National Taiwan University} 
}

\begin{document}
\maketitle

\renewcommand{\thefootnote}{\fnsymbol{footnote}}
\footnotetext[3]{Equal contribution}

\begin{abstract}
In the context of autonomous navigation, effectively conveying abstract navigational cues to agents in dynamic environments presents significant challenges, particularly when navigation information is derived from diverse modalities such as both vision and high-level language descriptions. To address this issue, we introduce a novel technique termed `Virtual Guidance,' which is designed to visually represent non-visual instructional signals. These visual cues are overlaid onto the agent's camera view and served as comprehensible navigational guidance signals. To validate the concept of virtual guidance, we propose a sim-to-real framework that enables the transfer of the trained policy from simulated environments to real world, ensuring the adaptability of virtual guidance in practical scenarios. We evaluate and compare the proposed method against a non-visual guidance baseline through detailed experiments in simulation. The experimental results demonstrate that the proposed virtual guidance approach outperforms the baseline methods across multiple scenarios and offers clear evidence of its effectiveness in autonomous navigation tasks.
\end{abstract}    
\section{Introduction}
\label{sec:intro}
Biological instincts are highly attuned to visual cues, a capability that is of paramount importance for executing navigation tasks effectively. For instance, Google Maps employs a combination of arrows and augmented reality to guide users to their desired destination. In contrast, navigational guidelines presented in textual or numerical formats, such as `turn 15 degrees at the next building’ or `proceed for 500 meters,’ are less straightforward and necessitate the high-level abstract information to be encoded into a form that can more easily be comprehended. Although certain studies have endeavored to transform these abstract navigational instructions into intermediate guidance signals (e.g., locational waypoints) that provide agents with spatial or directional cues~\cite{Muller2018CORL, bansal2019-lb-wayptnav, Krantz_2021_ICCV}, these often remain counterintuitive for model-free agents to easily grasp the environmental dynamics for effective navigation.  This inadequacy becomes particularly pronounced when considering environments characterized by dynamic variables, such as moving objects, uncertainties, or other evolving conditions, where mere reliance on spatial or directional cues is inadequate. Moreover, the integration of disparate modalities (e.g., image observations and language instructions) remains non-trivial due to the inherent challenges in concurrently reconciling the information derived from different modalities.

\begin{figure}[t]
  \centering
  \includegraphics[width=\linewidth]{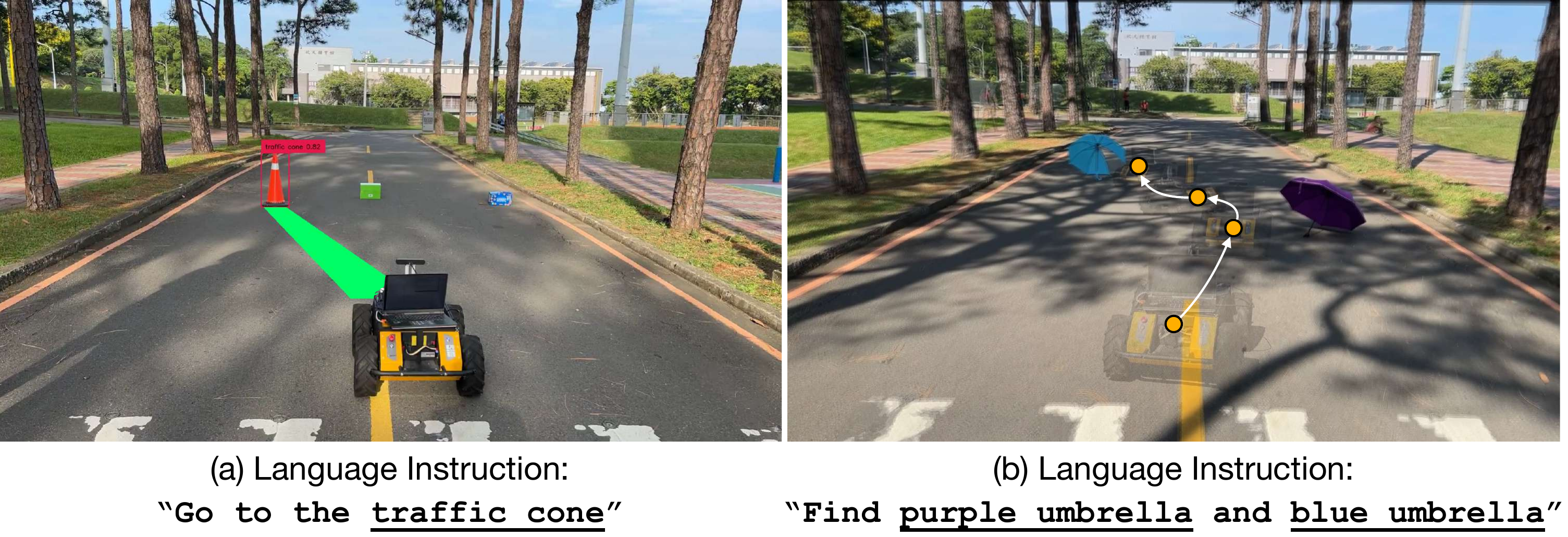}
  \caption{Demonstration of (a) the concept of virtual guidance, and (b) the agent’s behaviors in $\{\trafficcone, \greenbox, \bluebox\}$ and $\{\blueumbrella, \purpleumbrella\}$ scenarios.}
  \label{fig:teaser}
\end{figure}

Despite the aforementioned challenges, there is still often a necessity for autonomous agents in visual navigation tasks to manage multiple modalities. These modalities are essential for both local control mechanisms and long-term planning strategies. Many previous navigation tasks have employed hierarchical frameworks to handle these diverse modalities. For example, the planning component may rely on non-visual cues such as LiDAR, GPS, or text instructions, while the local controller utilizes visual observations. Recent years have seen considerable success in integrating instructions into robot navigation tasks through such hierarchical methods~\cite{Guldenring2020IROS, Pokle2019DeepLT, LiCoRL2019, Faust2017PRMRL, Linh2021IROS, LinhIROS2021-2, Jan2021ICRA, Brito2021RAL, Linh2023SII, Chaplot_2020_CVPR}. However, such approaches necessitate that the agent first learns a mapping mechanism to correlate navigation information with observed images. In light of the extensive research on various deep neural network models and their demonstrated efficacy in identifying complex interrelationships within an image, a potential promising avenue in the realm of visual navigation could lie in the direct representation of navigation instructions on visual perceptions. This can potentially alleviate the cognitive burden on agents by rendering non-visual, abstract modalities more easily comprehensible.

Inspired by~\cite{Mueller2018CoRL, Sax2020CoRL, Chen2020CoRL, Yang2022IROS, Hong2018IJCAI}, we propose a method of representation for non-visual instructional or modal signals as visual format, which we term `\textit{virtual guidance}.' The virtual guidance signals, rendered as either colored paths or spheres, are superimposed on the semantic segmentation derived from the agent's camera view, with the goal of guiding the agent toward a specific direction, as illustrated in Fig.~\ref{fig:teaser}~(a). This concept bears similarities to technologies such as Google Maps' Line View, which incorporates Augmented Reality (AR) guidance based on Global Positioning Systems for intuitively guiding human, while this work aims to extend the concept to incorporate the vision-based guidance signals to guide deep reinforcement learning (DRL) agents. Our objective is to assess the efficacy of vision-based virtual guidance schemes for enhancing agent navigation capabilities over various environmental contexts.

To implement and evaluate the effectiveness of virtual guidance schemes, we developed a sim-to-real framework that enables a DRL agent to be trained in simulation with virtual guidance signals and then deployed in a real-world environment without the need for fine-tuning. For the evaluation in simulation, we developed multiple environments for experiments using Unity~\cite{unity-eng} to support the rendering of virtual guidance signals, which are not naively supported by existing off-the-shelf simulation platforms~\cite{habitatchallenge2023}. For real-world validation, we extend this concept by transforming language instructions into virtual guidance signals with the concept of AR. Specifically, we employ a Large Language Model (LLM) with few-shot prompt engineering to interpret language instructions into sub-goals, and render them into corresponding virtual guidance signals with AR to guide the DRL agents. This process is illustrated in Fig.~\ref{fig:teaser}. To ensure accurate and consistent placement of virtual guidance over time, we train a scene coordinate regression (SCR) model for visual re-localization and introduce a two-stage process to position virtual objects using pseudo scene coordinates. This ensures that waypoints corresponding to sub-goals are correctly positioned relative to the scene geometry. Through extensive experiments, we observe that integrating guidance signals directly into the agent's observations enhances its performance in adhering to designated routes, and outperforms the baseline approach that relies solely on non-visual instructions. The primary contributions of this paper can be summarized as follows:
\begin{itemize}
\item We introduce the concept of virtual guidance as a type of mid-level representation, and validate its adaptability and feasibility for transferring policies pre-trained in simulated scenarios to the real world without fine-tuning.
\item We extend the virtual guidance concept into real-world scenarios by integrating LLM and an open-vocabulary object detection model for guidance generation. This approach transforms language instructions into visual cues for agents to comprehend and follow. This demonstrates the potential of navigation based on flexible instructions.
\item We introduce a two-stage process for waypoint placement using pseudo scene coordinates, achieving more accurate positioning than those with estimated coordinates.
\item We provide a comprehensive set of analyses and demonstrate that rendering guidance signals directly onto the agent's observations results in superior performance as compared to non-visual guidance in navigation tasks.
\end{itemize}

\section{Related Work and Background}
\label{sec::related_work}

\subsection{Navigation and Guidance Mechanisms}
Several navigation frameworks have been developed to direct an agent toward specified destinations. One common branch involves leveraging hierarchical frameworks to transmit navigational instructions or information to the agent~\cite{Guldenring2020IROS, Pokle2019DeepLT, LiCoRL2019, Faust2017PRMRL, Linh2021IROS, LinhIROS2021-2, Jan2021ICRA, Brito2021RAL, Linh2023SII, Chaplot_2020_CVPR}. Such non-vision-based guidance mechanisms can broadly be categorized as (1) implicit guidance and (2) vector-based guidance. The former offers the agent spatial knowledge through environmental cues, such as a localized map~\cite{Gupta2017CVPR, Chaplot_2020_CVPR, Liang2021ICRA} or a set of waypoints~\cite{Bansal2020CoRL, Linh2021IROS}. In contrast, the latter instructs the agent on specific actions or orientations to adopt~\cite{Linh2023SII}. To elaborate on this latter category, previous research endeavors have experimented with giving agents directional guidance toward targets or waypoints using non-visual mechanisms~\cite{Gao2017CoRL, Linh2023SII, Mousavian2019ICRA, Brito2021RAL}. Beyond these two categories, recent literature has also delved into object-based navigation, where the agent is guided by images of the target object or area~\cite{Zhu2017ICRA}. Some other research efforts represent waypoints as a sequence and instruct the agent to reach them in an ordered fashion~\cite{Gupta2017CVPR, Gao2017CoRL, Mousavian2019ICRA, Fang2019CVPR, Bansal2020CoRL, Chaplot2020NeurIPS, Liang2021ICRA, Ramakrishnan2022CVPR, Chaplot_2020_CVPR}, or through text-based instructions in vision-language frameworks. In this paper, the focus lies on explicit guidance realized through visual rendering of guidance signals. Unlike previous methodologies that concatenate observations with various modalities, our approach incorporates path or waypoint guidance directly into the agent's observations.

\begin{figure*}[t]
  \centering
  \includegraphics[width=.97\textwidth]{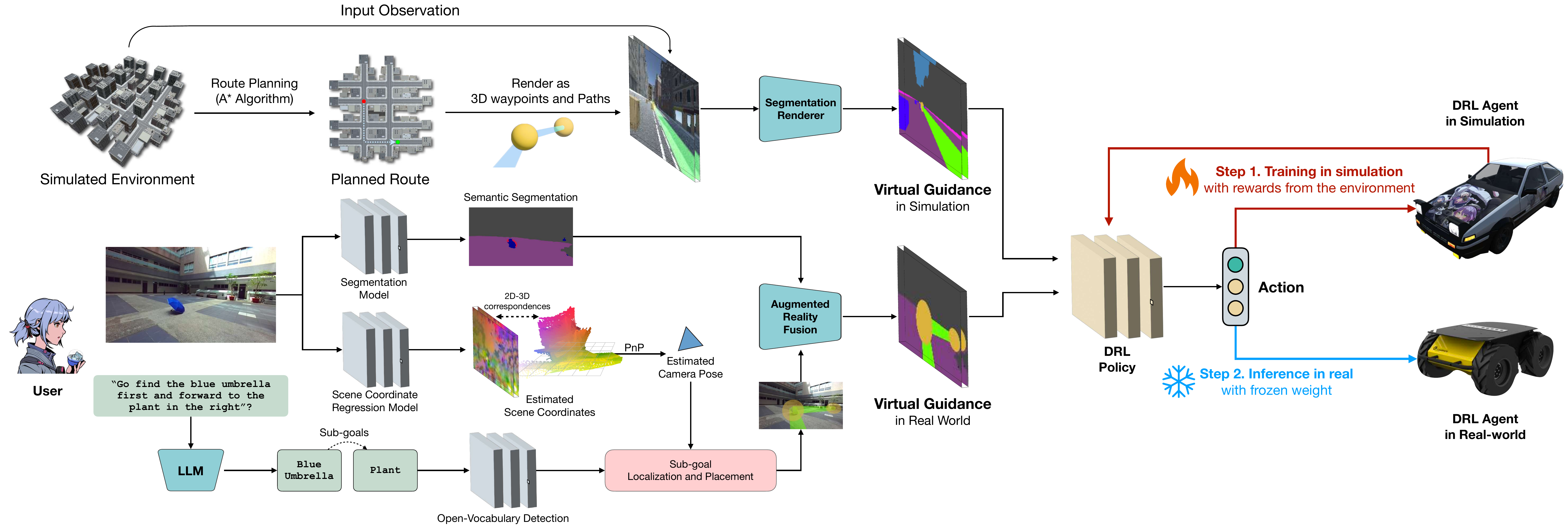}
  \caption{The illustration depicts the proposed simulation-to-real transfer framework for navigation tasks with virtual guidance.  The color set for rendering is aligned to bridge the domain gaps between simulated and real observations from the DRL agents.}
  \label{fig:overview}
\end{figure*}

\subsection{Mid-Level Representation based Navigation}
To implement virtual guidance, this paper explores approaches for the representation of virtual guidance signals based on mid-level representations. Mid-level representations are abstract concepts that capture physical or semantic meanings, and are typically domain-invariant properties extracted from visual scenes. Such representations have found applications in robotics for conveying information from perception modules to control modules~\cite{Hong2018IJCAI, zhao2020sim2real}. These mid-level representations can assume various forms, such as depth maps, optical flow, and semantic segmentation, each possessing unique strengths and weaknesses in different scenarios~\cite{Yang2022IROS}. A comprehensive, expressive, and interpretable mid-level representation is essential for the success of modular, learning-based frameworks. Existing research on navigation based on mid-level representations has primarily focused on facilitating obstacle avoidance or random path following for robotic agents, often in the absence of explicit instructional or guidance signals on direction or path~\cite{Hong2018IJCAI, Yang2022IROS}. This study extends this domain by introducing virtual guidance as mid-level representation.

\subsection{Reinforcement Learning and Soft Actor-Critic}
In DRL, an agent interacts with an environment $\mathcal{E}$ characterized by Markov Decision Process (MDP), by observing a state $s_t\in\mathcal{S}$ from the state space $\mathcal{S}$ at timestep $t$, executing an action $a_t\in\mathcal{A}$ selected from an action space $\mathcal{A}$ based on a policy $\pi$, and receiving a reward $r_t$ from $\mathcal{E}$. The aim of the agent is to learn an optimal $\pi^*$ that maximizes the expected return $G_t=\mathbb{E}\big[ \sum^{T}_{\tau = t} \gamma^{\tau-t} r_\tau \big]$, where $\gamma$ is a discount factor, and $T$ is the time horizon of an episode~\cite{intro2rl,sutton1999between}. To enhance the explorative behaviors of an RL agent, maximum entropy RL~\cite{energy-based-rl} proposes to find a $\pi$ that maximizes the expected return along with the entropy of $\pi$, given by $G_t=\mathbb{E}[\sum^T_{\tau=t}\gamma^{\tau-t}(r_\tau + \alpha\mathcal{H}(\pi))]$, where $\mathcal{H}(\pi)\triangleq\mathbb{E}[-\log\pi(\cdot|s_t)]$ denotes the entropy of $\pi$, and $\alpha$ is a temperature parameter, which controls the contribution of the entropy term to $G_t$. The Soft Actor-Critic (SAC)~\cite{discrete-sac, sac2, hybrid-control} algorithm further introduces the maximum entropy principles into deep reinforcement learning, which expands the capabilities of RL through the utilization of deep neural networks for handling complex, high-dimensional state spaces.
SAC integrates a parameterized soft-Q function $Q_\theta$ and a stochastic policy $\pi_\phi$, where $\theta$ and $\phi$ denote the weights of DNNs. SAC facilitates advanced exploration and performance in various control scenarios.

\begin{figure*}[t]
  \centering
  \includegraphics[width=\textwidth]{images/figure3-virtual_guidance_example.pdf}
  \caption{\textbf{Left (a)-(e):} An overview of different types of virtual guidance schemes compared to the vector based one; \textbf{Right (f):} The illustration of two types of planning schemes considered in the simulated environments for generating virtual guidance.}
  \label{fig:virtual_guidance}
\end{figure*}

\section{Virtual Guidance for Navigation with Simulation and Augmented Reality}
\label{sec::methodology}
In this work, we introduce a comprehensive framework for virtual guidance generation across both simulated and real-world environments, coupled with a sim-to-real training methodology for effective policy transfer. The overview of the framework is illustrated in Fig.~\ref{fig:overview}. In simulation, the virtual guidance is rendered as semantic segmentation and serves as DRL agents' input observation. The utilization of semantic segmentation facilitates the transfer from virtual to real-world scenarios~\cite{Hong2018IJCAI, zhao2020sim2real}, which allows images from both domains to be converted into a uniform representation and serve as inputs for the DRL agent. Such a training strategy offers a viable solution for minimizing the perception of domain discrepancies by the DRL agent, and thus enhances its adaptability and performance following the transfer.
On the other hand, in real-world scenarios, the learned policy is frozen and transferred from simulation. To realize the virtual guidance in real world, the virtual guidance is rendered through augmented reality (AR) that enables seamless integration of virtual elements into real environments. Specifically, a SCR is applied to perform camera pose estimation, and the estimated camera pose as well as the scene coordinates are used for re-localization and for rendering virtual path from specific camera view. To demonstrate the flexibility of the transferred policy and virtual guidance in downstream applications, we further integrate an LLM for parsing natural language directives into navigational sub-goals. This transforms verbal instructions into spatial waypoints that enable navigation based on language inputs.

\subsection{Virtual Guidance in Simulated Environments}
\label{subsec::simulation-methodology}
To investigate the effectiveness of virtual guidance as a form to guide DRL agents, we developed a flexible framework using the Unity engine~\cite{unity-eng} and the Unity ML-Agents Toolkit~\cite{ml-agents}, as illustrated in Fig.~\ref{fig:overview}, This framework is designed to generate configurable virtual guidance signals, including different navigation paths and a range of training and evaluation scenarios. This design enables the exploration of diverse ways of presenting guidance signals to the agent, such as various virtual guidance or vector-based schemes. The agent receives the guidance signals along with semantic segmentation maps, and is tasked with processing them to learn a policy to reach its destination.

\subsubsection{Virtual Guidance Generation in Simulation}
\label{subsubsec:virtual-guidance-generation}
This section presents the workflow for generating virtual guidance in simulation for validating the proposed methodology, which consists of two components: (a) a \textit{planning module} responsible for determining the navigation trajectory, and (b) a \textit{virtual guidance representation module} tasked with rendering the virtual guidance for the agent.

\subsubsection{Planning Module} 
\label{subsubsec:virtual-guidance-planning}
Fig.~\ref{fig:virtual_guidance}~(f) depicts the possible planning schemes of the planning module. The navigation trajectories can be planned between any two points, whether from the starting point $S$ to the destination $D$, agent's position $P_t$ to waypoints $\mathcal{W}$, or $P_t$ to $D$, with the A* algorithm~\cite{AStarAlgorithm} employed in this study. In the simulated environments, it is assumed that the planning module has awareness of agent's position $P_t$ to mitigate any potential errors in the localization process, while in the real-world scenario, $P_t$ is estimated by a scene coordinate regression model to perform visual re-localization.

The planned trajectory can be either rendered as a vision-based virtual guidance signal or transformed into vectors for the agent's interpretation. This trajectory can be generated in (a) \textit{one-time} mode at the beginning of the episode from $S$ to $D$ through several waypoints $\mathcal{W}$, or (b) \textit{real-time} mode from either $P_t$ to $\mathcal{W}$ or $P_t$ to $D$. In the simulation, 
we investigate both \textit{one-time} and \textit{real-time} configurations for trajectory generation, and provide analyses in Section~\ref{sec:comparison-of-vg}.

\subsubsection{Virtual Guidance Representation Module} 
\label{subsubsec:virtual-guidance-representation}
Once the navigation trajectory is obtained from the planning module, virtual guidance can be generated and rendered on semantic segmentation to enable the agent to recognize its meaning. The proposed representation schemes are illustrated in Fig.~\ref{fig:virtual_guidance} and are elaborated as follows.

\paragraph{Navigation Path.} In this scheme, the navigation line obtained from the planning module is represented as a colored path on the semantic segmentation map. An example visualization is illustrated in Fig.~\ref{fig:virtual_guidance}~(c). Specifically, the navigation path is implemented as a 3D mesh and projected onto the camera view plane. This rendered navigation path can be considered as a rich and informative signal that carries both semantic and guidance information. 

\paragraph{Waypoint.} The second scheme generates a set of waypoints $\mathcal{W}$ by segmenting the planned navigation trajectory, where different waypoints are spaced with a regular distance from $S$ to $D$. The waypoints serve as hints to instruct the agent to the destination. These waypoints are visualized as 3D virtual balls in the virtual environments and are projected onto the camera image plane. The visualization is presented in Fig.~\ref{fig:virtual_guidance}~(d). Unlike the first scheme, which utilizes a navigation path to provide dense and informative signals, the second one provides the waypoints as 3D virtual balls, which are sparse signals for the DRL agent to locate.

\subsection{Virtual Guidance in Real-World Scenarios}
\label{sec::methodology-real}
In this section, we present the methodology for representing virtual guidance in real-world settings using text prompts and high-level natural language descriptions. This validation aims to further demonstrate the flexibility of our methodology, as it enables the generation of virtual guidance signals from more general guidance sources to direct our robot's actions. Our methodology enables seamless transfer to the real-world without additional fine-tuning through the use of mid-level representation, which enhances the adaptability and transferability of our method to practical scenarios, making it suited for real-world applications.

\begin{figure}[t]
  \centering
  \includegraphics[width=\linewidth]{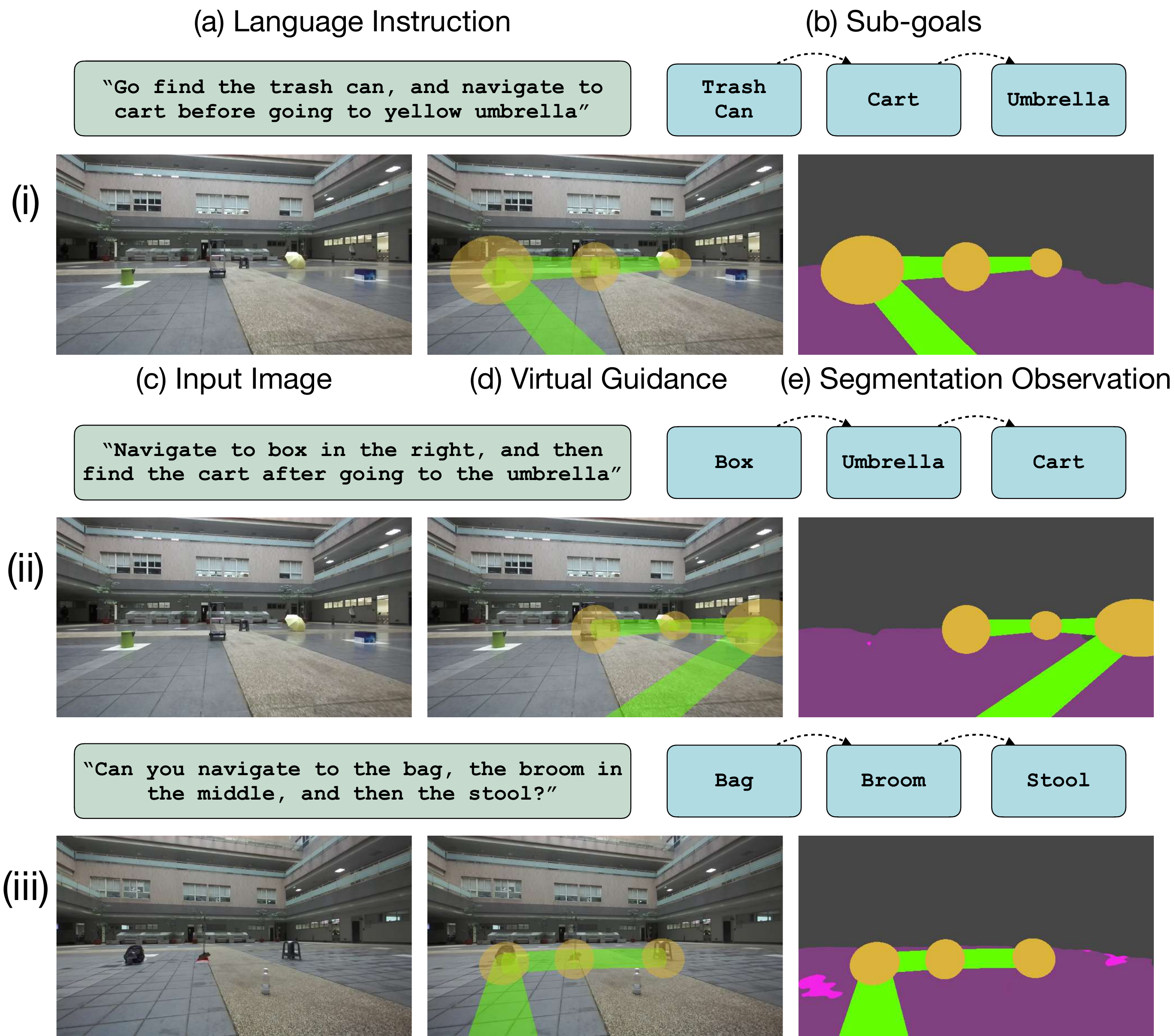}
  \caption{The illustrations of sub-goals generated from language instructions with their corresponding virtual guidance. Cases (i) and (ii) demonstrate scenarios with the same object placements but different instructions, while case (iii) presents another example.}
  \label{fig:subgoal-generation}
\end{figure}

\subsubsection{Sub-Goals and Target Generation with Open Set Object Detection and Large Language Model}
\label{sec::sub-goals}
To demonstrate the adaptability and flexibility of our framework in real-world settings, we provide a high-level description in human language, comprising a sequence of target objects with appearance descriptions. We then employ GPT-4~\cite{gpt-4}, a LLM preconfigured with few-shot prompting engineering, to interpret these high-level descriptions and generate several sub-goals. Each sub-goal contains text describing the appearance of the target object towards which the robot is to navigate. Fig.~\ref{fig:subgoal-generation} presents examples of language instructions and their corresponding generated sub-goals. These generated sub-goals are sequentially fed into YOLO-World~\cite{cheng2024yolow}, an open-vocabulary object detection model capable of zero-shot object detection, to produce a sequence of detections according to the provided language prompts.

\subsubsection{Virtual Guidance Generation in Real with Visual Relocalization and Augmented Reality}
With the established sub-goals, the next objective is to generate virtual guidance signals in the real-world. The target objects identified by the sub-goals are then processed through the virtual guidance representation module, which consists of two phases: (1) visual re-localization and (2) waypoint position retrieval through 2D-3D correspondence.

To render augmented reality objects, it is crucial to re-localize the camera's position and orientation to ensure consistent virtual guidance over time. In our work, we train a regression model to estimate the scene coordinates, denoted by $\mathcal{Y} = \{\mathbf{\hat{y}}_i\}_{i=1}^N$, where each estimated scene coordinate  $\mathbf{\hat{y}_i}$ represents the 3D position in the world coordinate system corresponding to the pixel at index $i$. Following~\cite{brachmann2023ace}, we collect a set of $M$ training images $\{\mathcal{I}_j\}_{j=1}^M$ from the target scene and perform Structure-from-Motion (SfM)~\cite{schoenberger2016sfm}, where each image $\mathcal{I}_j \in \mathbb{R}^{H \times W \times 3}$ is associated with a pseudo camera pose $\mathbf{h^*_j}$. The projection error $r$ is applied to optimize the SCR model and is defined as the following:
\begin{equation}
    r(\mathbf{p_i}, \mathbf{\hat{y}_i}, \mathbf{h^*}) = \left\| \mathbf{p_i} - \mathbf{K}\, \mathbf{h^{*-1}}\, \mathbf{\hat{y}_i} \right\|,
\label{eq:reproject}
\end{equation}
where $\mathbf{p_i}$ denotes the observed pixel position, $\mathbf{\hat{y}_i} \in \mathcal{Y}$ represents the estimated 3D scene coordinate corresponding to pixel $\mathbf{p_i}$, $\mathbf{K}$ is the intrinsic matrix of the camera, and $\mathbf{h^*}$ is the pseudo camera pose. This error quantifies the discrepancy between the observed pixel position and the projected location of the 3D point, and its minimization refines the camera pose by aligning the projected 3D points with their corresponding observed 2D positions. To infer the camera pose $\mathbf{h}$ from the estimated scene coordinates, we employ the Perspective-n-Point (PnP) algorithm, which computes the estimated camera pose $\mathbf{h}$ by solving for the transformation that best maps the 2D-3D correspondences.

\begin{figure}[t]
  \centering
  \includegraphics[width=\linewidth]{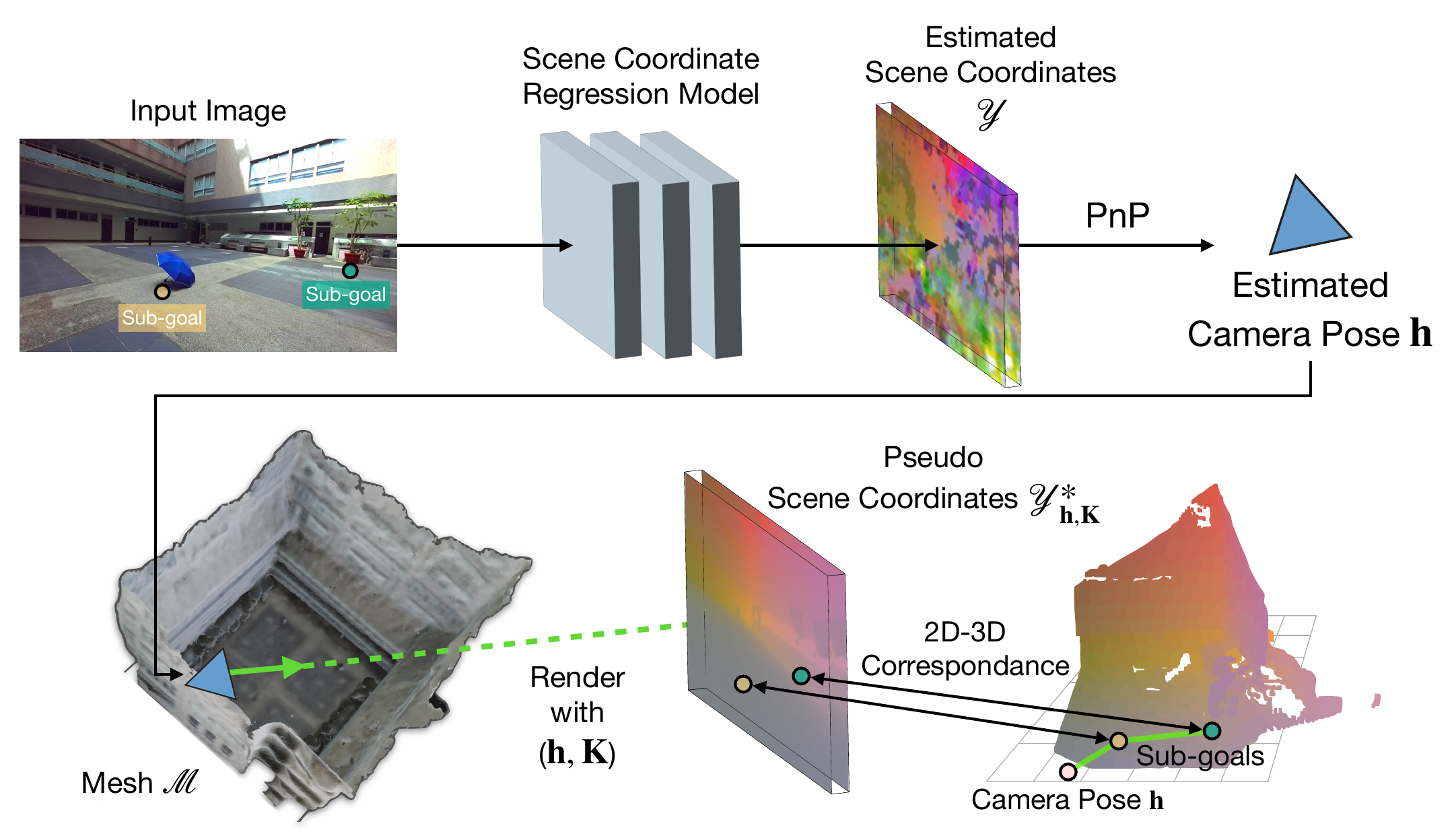}
  \caption{The illustration of the proposed two-stage process to retrieve sub-goal positions through pseudo scene coordinates.}
  \label{fig:psuedo-scene-coordinates}
\end{figure}

To place the waypoints (i.e., sub-goals) and the pathways towards these sub-goals into 3D positions for the preparation of AR visualization, we utilized scene coordinates for mapping pixel coordinates into world coordinates. These scene coordinates can either be referred from the estimated scene coordinates $\mathcal{Y}$ or the pseudo scene coordinates $\mathcal{Y}^*$ from the reconstructed mesh. In this work, we use $\mathcal{Y}^*$ over $\mathcal{Y}$ because the accuracy of 2D-3D correspondences in $\mathcal{Y}$ varies across positions, despite an accurately estimated camera $\mathbf{h}$ being inferable from $\mathcal{Y}$. This limitation arises because the SCR model is primarily optimized with reprojection error, and the PnP approach requires only a few accurate correspondences for precise camera positioning. Consequently, we leverage the accurately estimated camera pose $\mathbf{h}$, inferred from $\mathcal{Y}$, and refer back to the reconstructed SfM model to generate $\mathcal{Y}^*$. This process is illustrated in Fig.~\ref{fig:psuedo-scene-coordinates}. Specifically, we extend the reconstructed SfM points to a dense triangulated mesh $\mathcal{M}$ using Poisson surface reconstruction~\cite{schoenberger2016sfm}. The pseudo scene coordinate $\mathcal{Y}^*$ can be obtained through a rendering process applied to the mesh:
\begin{equation}
\mathcal{Y}^*_{\mathbf{h}, \mathbf{K}} = \mathcal{R}(\mathcal{M}, \mathbf{h}, \mathbf{K}),
\end{equation}
where $\mathcal{Y}^*_{\mathbf{h}, \mathbf{K}}$ denotes the pseudo scene coordinate map rendered from mesh $\mathcal{M}$ under pose $\mathbf{h}$ and intrinsics $\mathbf{K}$, $\mathcal{R}$ denotes the rendering operation that projects the mesh $\mathcal{M}$ according to $\mathbf{h}$ and $\mathbf{K}$. For a pixel $\mathbf{p}_q \in \mathbb{R}^2$, its corresponding 3D position $\mathbf{y}_q \in \mathbb{R}^3$ is looked up based on the following:
\begin{equation}
\mathbf{y}_q = \mathcal{Y}^*_{\mathbf{h}, \mathbf{K}}(\mathbf{p}_q).
\end{equation}
We then process the detection boxes of all identified sub-goals with this look-up function, to retrieve the 3D locations of sub-goals and serve as waypoints. The waypoints and their connecting pathways are rendered as virtual guidance on the mid-level representation (i.e., semantic segmentation), serving as input observation for the DRL agent.

\subsubsection{Simulation-to-real Transfer for Virtual Guidance}
\label{sec::virtual-to-real-transfer}
In order to bridge the domain gap between virtual and real observations, the virtual guidance signals and the semantic segmentation, derived from the real world, are rendered and aligned using the same color set as those configured in the simulation environment. This enables the pre-trained DRL agent's policy to be transferred to the real world without the need for additional fine-tuning. Thus, the agent can leverage its learned knowledge from simulation to follow guidance and control the robot in real-world scenarios.  our experiments, the robotic platform utilized is a ClearPath Husky Unmanned Ground Vehicle, and the semantic segmentation model employed is Rainbow~\cite{rainbow} with twelve categories. 
\section{Experimental Results}
\label{sec::experimental_results}

\subsection{Experimental Setup}
\subsubsection{Simulation Environment Setup}
\label{subsubsec:virtual-env}
\begin{figure}[t]
  \centering
  \includegraphics[width=\linewidth]{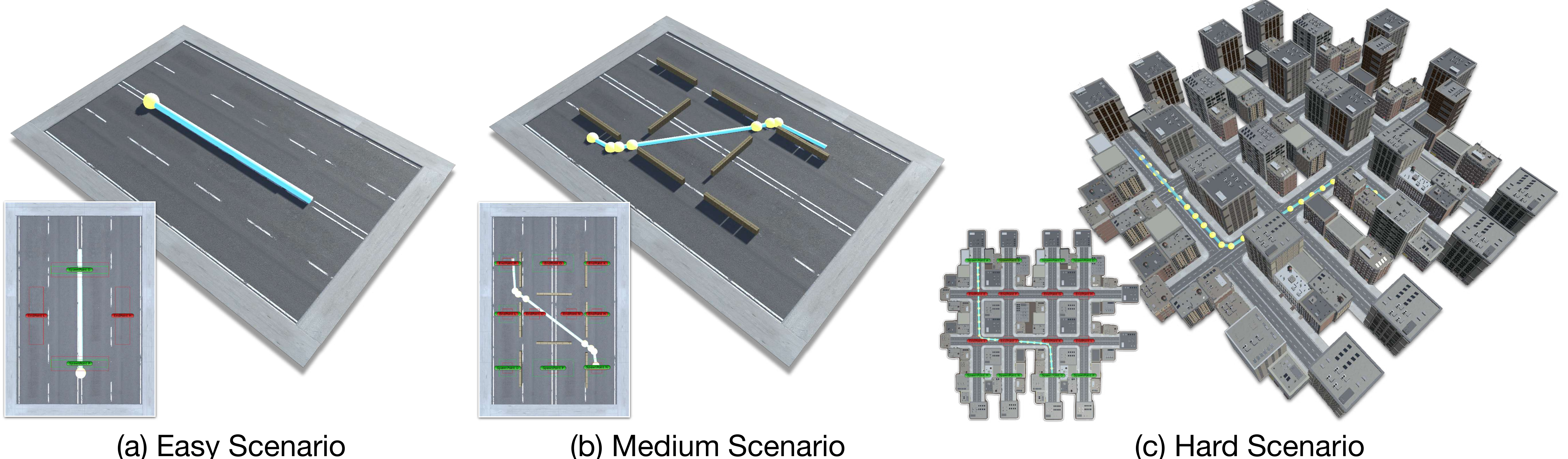}
  \caption{Simulation environments designed for experiments.}
  \label{fig:map}
\end{figure}

To evaluate the efficacy of virtual guidance, we developed three simulation environments using the Unity engine~\cite{unity-eng}. These environments vary difficulty levels: \textit{Easy}, \textit{Medium}, and \textit{Hard}, each designed to meet distinct evaluation purposes. The \textit{Easy} scenario comprises a simple guidance path featuring only one waypoint (i.e., the destination) to assess the agents' capability to adhere to straightforward virtual instructions. The \textit{Medium} scenario expands upon this by extending into multiple waypoints and the path segments, thereby increasing the distance to the destinations. The \textit{Hard} scenario is designed to simulate an urban landscape with eight intersections. In this scenario, a set of 89 routes are designed for the training phase; while during the evaluation phase, the agent is evaluated under four unseen routes to examine its capacity for navigating in unfamiliar routes with virtual guidance. The illustrations of these scenarios are presented in Fig.~\ref{fig:map}. In particular, the \textit{Medium} and \textit{Hard} scenarios introduce obstacles that the agent is required to avoid while adhering to the provided guidance signals.

\subsubsection{Agent Setup}
In our experiments, the DRL agent is implemented with a DNN and is trained with the SAC algorithm~\cite{sac1, discrete-sac}. The agent's observation space consists of three stacked semantic segmentation frames, each having dimensions of $84 \times 180$. These frames can be rendered either with or without virtual guidance. The agent maintains a consistent velocity $v$ of $\SI{6}{\meter/\second}$ while navigating, with its action space includes two primary actions: (a) maintaining the agent's current directional orientation, or (b) incremental adjusting the orientation with $\alpha$ angle. The sign of $\alpha$ is essential: negative values result in a leftward adjustment, while positive values prompt a rightward shift. This mechanism endows the agent with the ability to navigate and execute turns in a non-binary manner. The angular velocity $\omega$, influenced by these continuous adjustments to $\alpha$, is formulated as the following:
\begin{equation}
    \omega \pluseq \alpha \times \kappa \times \Delta t,
\end{equation}
where $\Delta t$ represents the time interval, and $\kappa$ is the steering sensitivity. We set $\alpha$ to a standard value of either $-\SI{35}{\degree/\second^2}$ or $\SI{35}{\degree/\second^2}$ and $\kappa$ to two. Rather than relying on abrupt and binary changes in direction for the agent's action, the agent's action is decided based on the cumulative effects of successive adjustments to $\alpha$.

\begin{figure}[t]
  \centering
  \includegraphics[width=\linewidth]{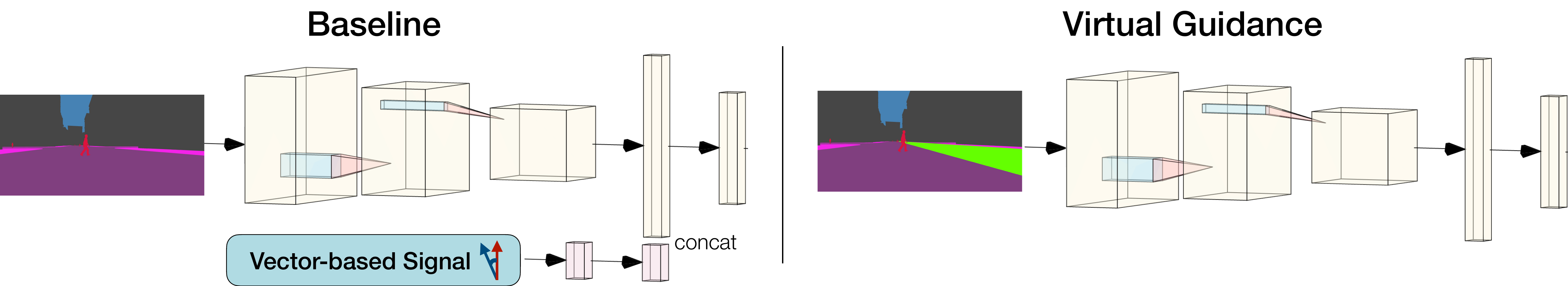}
  \caption{Comparison between baseline and virtual guidance.}
  \label{fig:baseline-real-vg}
\end{figure}

\subsubsection{Baseline}
\label{subsec:baseline}
To evaluate the effectiveness of the vision-based virtual guidance representation schemes described in Section~\ref{sec::methodology}, we introduce a baseline approach (denoted as `\textit{SegVec}'), which combines stacked segmentation observations with vector-based guidance as the agent's inputs. An example of the baseline input is illustrated in Fig.~\ref{fig:virtual_guidance}~(e), and a comparison of the architectures between the baseline and the proposed virtual guidance is presented in Fig.~\ref{fig:baseline-real-vg}. In the \textit{SegVec} baseline, two parameters are provided to the agent: the distance $r$ to the closest forthcoming waypoint $\omega^*$ and the agent's orientation $\Theta$ with respect to $\omega^*$. These are presented in polar coordinates as $(r, \Theta)$. The orientation $\Theta$ falls within the range of $-180$ to $180$ degrees, and is normalized to a range of $[-1, 1]$ before being provided to the navigation agent. 

\subsubsection{Reward Function}
During training, the reward combines navigation reward $R_{nav}$ and episodic reward $R_{goal}$. $R_{nav}$ encourages alignment with guidance while $R_{goal}$ reflects episode outcomes by signaling success with a positive reward or failure with a penalty in reaching the destination. Specifically, the reward for the agents trained with virtual guidance is formulated as:
\begin{equation}
R_{nav} = \begin{cases}
R_w & \text{if waypoint collected}, \\
R_p & \text{if on navigation path}, \\
\end{cases}
\end{equation}
\begin{equation}
R_p = \text{clamp}(\frac{W_p}{2} - d, R_{min}, R_{max}),
\end{equation}
where $d$ is the shortest distance from agent position $P_t$ to the navigation line, and $W_p$ is the width of visible navigation path. $R_{w}$ is a one-time reward awarded only when the agent collects a waypoint. $R_{p}$ is allocated to the agent at fix time duration, with the parameters $\{W_p, R_{min}, R_{max}, R_w\}$ set to $\{0.18, -0.1, 0.08, 5.0\}$ for \textit{Easy} and \textit{Medium} scenarios, and $\{1.2, -0.2, 0.4, 5.0\}$ for \textit{Hard} scenario. For $R_{goal}$, the agent receives $10.0$ when reaching the destination, while a penalty of $-10.0$ is applied if it collides with any obstacle, moves outside the boundaries, or exceeds the time. The final reward function $R$ is calculated as $R = R_{nav} + R_{goal}$. 

\begin{table}[t]
\caption{A comparison between the proposed virtual guidance and non-visual guidance in simulation under three scenarios.}
\label{tables:impact-analysis-simulation}
\centering
\resizebox{\linewidth}{!}{%
\renewcommand{\arraystretch}{1.7}
\begin{tabular}{ c | c | c | ccc }
\toprule
\multirow{2}{*}{\textbf{Scenario}}& \textbf{Guidance} & \textbf{Representation} & \multicolumn{3}{c}{\textbf{Performance}} \\
& \textbf{Scheme} & \textbf{Form} & \textit{SPL}~($\uparrow$) & \textit{Success Rate}~($\uparrow$) & \textit{Waypoint Collecting Rate}~($\uparrow$) \\
\midrule
Single Waypoint & $\textit{SegVec}$ & \{RGB, $(r, \theta)$\} & $\SI{88.8 +- 3.7}{\percent}$ & {$\SI{96.4 +- 3.4}{\percent}$} & $\SI{96.4 +- 3.4}{\percent}$\\
(\textit{Easy}) & $\textit{SegVisualGuide}$ & RGB & \textbf{${\SI{98.3 +- 0.5}{\percent}}$}  & \textbf{${\SI{99.3 +- 0.2}{\percent}}$} & \textbf{${\SI{99.3 +- 0.2}{\percent}}$} \\
\midrule
Multiple Waypoint & $\textit{SegVec}$ & \{RGB, $(r, \theta)$\} & $\SI{70.0 +- 2.7}{\percent}$  & $\SI{82.2 +- 9.2}{\percent}$   & \textbf{$\SI{89.3 +- 1.9}{\percent}$} \\
(\textit{Medium}) & $\textit{SegVisualGuide}$ & RGB & \textbf{$\SI{82.7 +- 0.9}{\percent}$}  & \textbf{$\SI{83.7 +- 0.8}{\percent}$} & $\SI{88.0 +- 2.3}{\percent}$ \\
\midrule
Multiple Waypoint & $\textit{SegVec}$ & \{RGB, $(r, \theta)$\} & $\SI{48.9 +- 7.7}{\percent}$ & $\SI{52.1 +- 2.0}{\percent}$  & $\SI{44.7 +- 2.0}{\percent}$   \\
(\textit{Hard}) & $\textit{SegVisualGuide}$ & RGB & \textbf{$\SI{67.5 +- 4.2}{\percent}$} & \textbf{$\SI{67.8 +- 4.2}{\percent}$} & \textbf{$\SI{79.3 +- 1.9}{\percent}$} \\
\bottomrule
\end{tabular}}
\end{table}
\begin{table}[t]
\caption{Analysis of failure cases in \textit{Medium} and \textit{Hard} scenarios.} 
\label{tables:failure-cases-comparison}
\centering
\resizebox{\linewidth}{!}{%
\renewcommand{\arraystretch}{1.4}
\begin{tabular}{ l | c | cc }
\toprule
~~\multirow{2}{*}{\textbf{Guidance Scheme}}~~ & \multirow{2}{*}{\textbf{Representation Form}} & \multicolumn{2}{c}{\textbf{Failure Cases}}\\
& & ~~~~\textit{Collision Rate}~($\downarrow$)~~~~ & ~~~~\textit{Out-of-Bound Rate}~($\downarrow$)~~~~ \\
\midrule
$\textit{SegVec}$ & \{RGB, $(r, \theta)$\} & $\SI{21.0}{\percent}$ & \bad{\textbf{$\SI{11.8}{\percent}$}} \\
$\textit{SegVisualGuide}$ & RGB & $\SI{21.6}{\percent}$ & $\SI{1.8}{\percent}$ \\
\bottomrule
\end{tabular}}
\vspace{-1em}
\end{table}

\subsubsection{Evaluation Metrics}
In our experiments, three metrics are utilized to evaluate the performance of the agents. These are described as follows.

\begin{enumerate}[label=\alph*)]
\item \textbf{Success rate:} 
This metric is utilized to assess the proficiency of the agent in reaching a designated destination.

\item \textbf{Success rate weighted by path length (SPL):} 
The SPL metric evaluates the agent's navigational performance by accounting for both the success in reaching the destination and the efficiency of the selected trajectory~\cite{SPL}: 
\begin{equation}
    SPL = \frac{1}{N} \sum_{i=1}^{N} S_i \cdot \frac{l_i}{\max(l_i, p_i)},
\end{equation}
where $S_i$ indicates whether the agent was successful in episode $i$, $l_i$ represents the shortest path distance from the agent’s starting position to the goal for episode $i$, and $p_i$ denotes the length of the path actually taken.

\item \textbf{Waypoints collecting rate:} It calculates the ratio between the number of waypoints collected and the total number of waypoints along the planned path.
\end{enumerate}

The reported results in simulation experiments are obtained by training the agent using three different random seeds, with each of them undergoing 1,000 inference runs.

\begin{figure}[t]
  \centering
  \includegraphics[width=\linewidth]{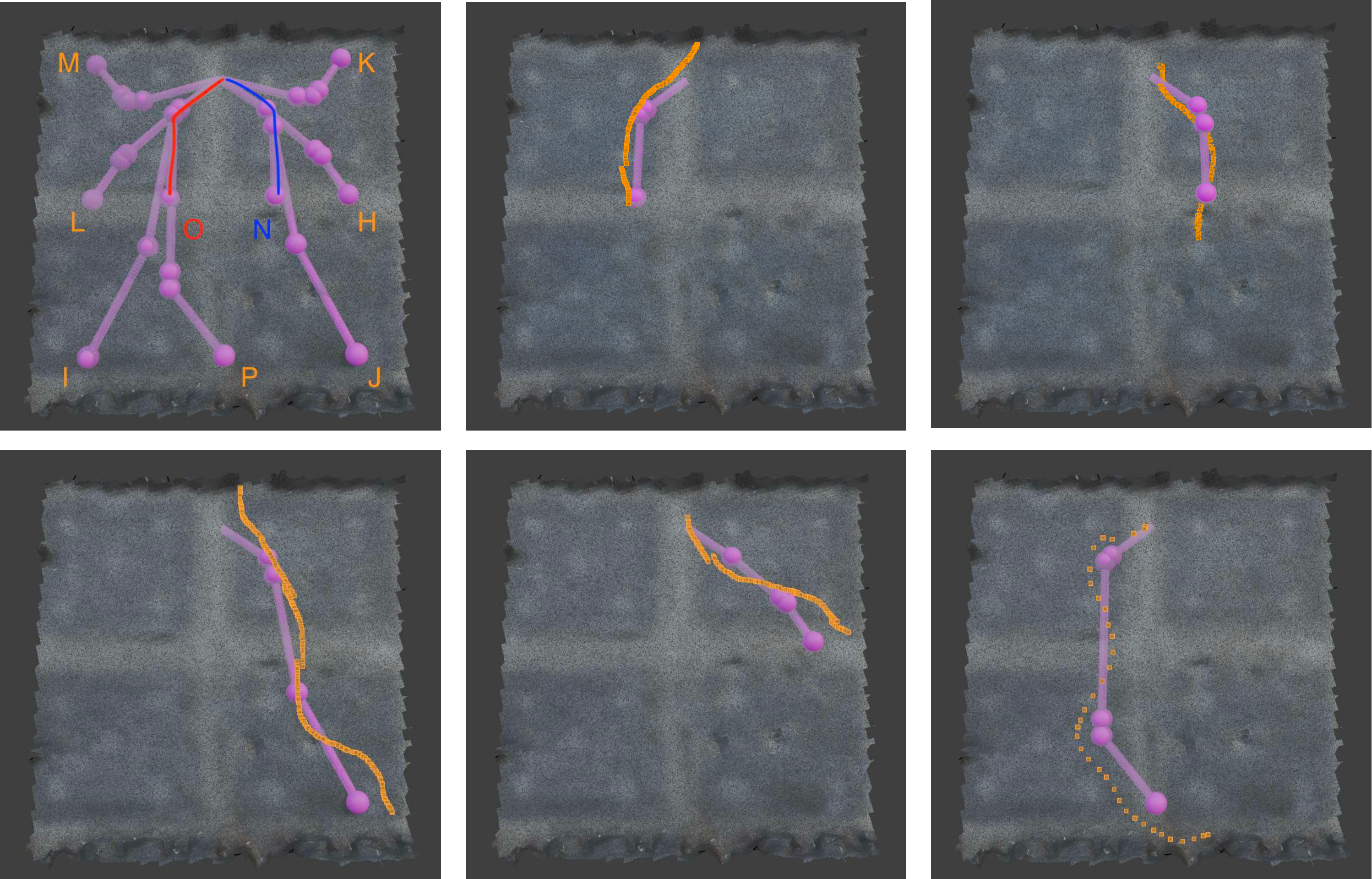}
  \caption{Illustration of the provided virtual guidance routes (denoted in pink) and the agents' navigation trajectories (denoted in orange) in preset routes scenarios in real-world experiments. The top-left figure shows the designed trajectories for evaluation.}
  \label{fig:real-trajectory}
\end{figure}

\subsection{Impact Analysis of Virtual Guidance}
\label{sec:impact-of-virtual-guidance}
This section presents a comparison between the proposed virtual guidance scheme and the baselines with non-visual guidance signals. The virtual guidance schemes were introduced in Section~\ref{subsubsec:virtual-guidance-representation}, while the baseline scheme was introduced in Section~\ref{subsec:baseline}. The evaluation results presented in Table~\ref{tables:impact-analysis-simulation} indicate that the schemes incorporating virtual guidance (denoted as `$\textit{SegVisualGuide}$') consistently outperform the baseline scheme (denoted as `$\textit{SegVec}$') in terms of SPL and success rate in all scenarios. To further evaluate the performance, we conduct an error analysis with the results presented in Table~\ref{tables:failure-cases-comparison}. The results indicate that agents trained with \textit{SegVec} exhibit a higher out-of-bound rate, which indicates an increased tendency to lost track from the planned navigation path. These observations reveal that the proposed vision-based guidance strategies can effectively offer informative navigational cues. They also suggest that virtual guidance can alleviate agent's burden to learn the correlation between visual observations and navigational instructions derived from non-visual modalities. 

\subsection{Real-World Validation of Virtual Guidance}
\label{sec:real-world-validation}
\begin{figure}[t]
  \centering
  \includegraphics[width=\linewidth]{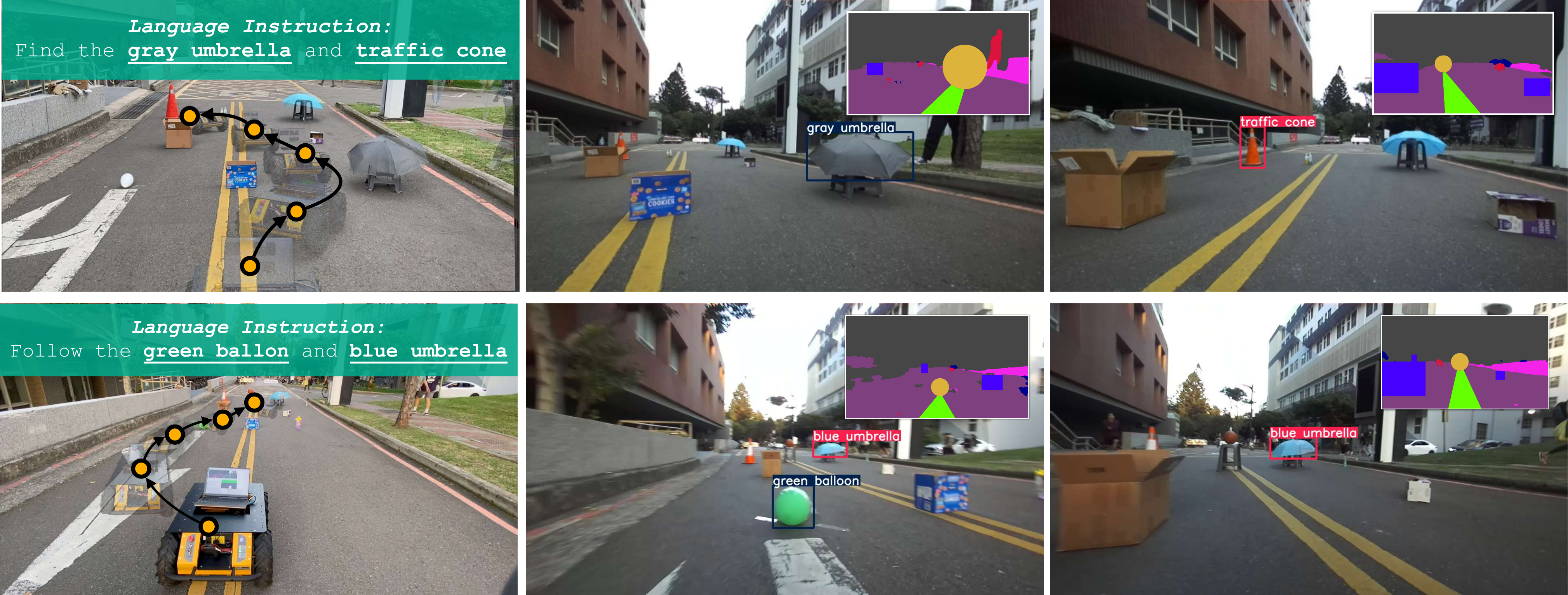}
  \caption{Visualizations of navigation trajectories with provided instructions and the corresponding generated virtual guidance.}
  \label{fig:real-demo}
\end{figure}

In this section, we explore and validate the application of the proposed virtual guidance concept in real-world contexts by transferring the DRL agent trained in simulation to the real world, as described in Section~\ref{sec::virtual-to-real-transfer}. In the real-world experiments, we design two scenarios for evaluation: (a) the preset routes scenario and (b) the language instruction scenario. In the preset routes scenario, we assess the agents' ability to follow predetermined navigation paths. On the other hand, in the language instruction scenario, we evaluate the agents' capability to navigate to specific sub-goals based on provided language instructions. The agent's task is to  approach the correct object(s) based on the provided text prompts and the transformed virtual guidance. Fig.~\ref{fig:real-trajectory} and~\ref{fig:real-demo} demonstrate the examples of these two scenarios respectively. Each result is obtained through thirty independent runs, and the results are presented in Table~\ref{tables:real-exp}. 

\begin{table}[t]
\caption{The real-world validation of the proposed virtual guidance (i.e., $\textit{SegVisualGuide}$) with the agents' policies transferred from simulation without further fine-tuning.}
\label{tables:real-exp}
\centering
\footnotesize
\resizebox{\linewidth}{!}{%
\renewcommand{\arraystretch}{1.2}
\begin{tabular}{ c | ccc }
\toprule
\rowcolor{lightgray!15} \multicolumn{4}{c}{\textbf{Preset Routes and Waypoints}} \\
\midrule
\multirow{2}{*}{\textbf{Scenarios}} & \multicolumn{3}{c}{\textbf{Performance}} \\
 & ~~~~~~~~\textit{SPL}~($\uparrow$)~~~~~~~ & ~~~~~~\textit{Success Rate}~($\uparrow$)~~~~~~ & ~~~~~~\textit{Waypoint Collecting}~($\uparrow$)~~~~~~ \\
\midrule
\textbf{\textit{Single Waypoint}}  & $\SI{64.0}{\percent}$  & $\SI{80.0}{\percent}$  & $\SI{80.0}{\percent}$ \\
\textbf{\textit{Multiple Waypoints}} & $\SI{57.0}{\percent}$  & $\SI{76.7}{\percent}$ & $\SI{82.0}{\percent}$ \\
\midrule
\rowcolor{lightgray!15} \multicolumn{4}{c}{\textbf{Language Instructions}} \\
\midrule
\textbf{\textit{Multiple Waypoints}} & $\SI{54.4}{\percent}$ & $\SI{73.3}{\percent}$ & $\SI{82.9}{\percent}$ \\
\bottomrule
\end{tabular}}
\end{table}
\begin{table}[t]
\caption{An comparison of different navigation guidance schemes.} 
\label{tables:impact-analysis-evaluation}
\centering
\resizebox{\linewidth}{!}{%
\renewcommand{\arraystretch}{1.7}
\begin{tabular}{ l | c | ccc }
\toprule
\multirow{2}{*}{\textbf{Guidance Scheme}} & \textbf{Representation} & \multicolumn{3}{c}{\textbf{Performance}} \\
& \textbf{Form} & \textit{SPL}~($\uparrow$) & \textit{Success Rate}~($\uparrow$) & \textit{Waypoint Collecting Rate}~($\uparrow$) \\
\midrule
$\textit{SegVec}$ & \{RGB, $(r, \theta)$\} & $\SI{48.9 +- 7.7}{\percent}$ & $\SI{52.1 +- 2.0}{\percent}$  & $\SI{44.7 +- 2.0}{\percent}$ \\
\midrule
$\textit{SegVisualGuide}$~($\textit{waypoint}$)  & RGB & {$\SI{57.7 +- 2.2}{\percent}$} & {$\SI{58.2 +- 2.0}{\percent}$} & $\SI{67.2 +- 1.3}{\percent}$ \\ 
$\textit{SegVisualGuide}$~($\textit{path}_{S\rightarrow D}$) & RGB & {$\SI{67.5 +- 4.2}{\percent}$} & {$\SI{67.8 +- 4.2}{\percent}$} & $\SI{79.3 +- 1.9}{\percent}$  \\ 
\vgPtoW & RGB & {$\SI{68.6 +- 2.7}{\percent}$} & {$\SI{69.1 +- 2.6}{\percent}$} & $\SI{78.9 +- 2.0}{\percent}$ \\ 
\bottomrule
\end{tabular}}
\end{table}

\begin{figure}[t]
  \centering
  \includegraphics[width=\linewidth]{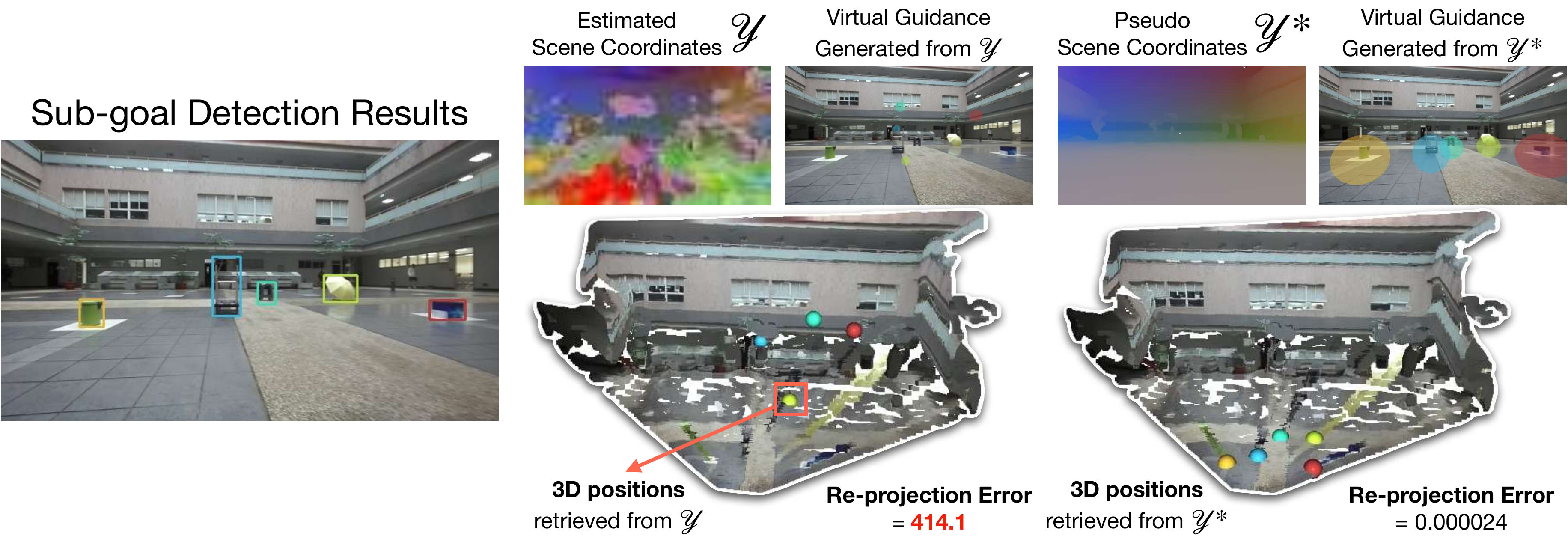}
  \caption{Comparison of waypoint positions obtained from estimated scene coordinates $\mathcal{Y}$ and pseudo scene coordinates $\mathcal{Y}^*$.}
  \label{fig:object-placement}
\end{figure}

These evaluation results demonstrate that the agents achieve a success rate comparable to that reported in Section~\ref{sec:impact-of-virtual-guidance} in real-world environments, indicating that the policies trained in virtual settings can be effectively transferred to real-world scenarios.  In the language instruction scenario, all objects described in the text prompts are correctly detected due to the capabilities of YOLO-World, and the agent is able to follow virtual guidance paths generated through language instructions. A few samples of the agents' navigation trajectories and the provided instructions are depicted in Fig.~\ref{fig:real-demo}. To validate the effectiveness of the proposed two-stage process for waypoint placement, we visualize the 3D positions obtained from both the estimated scene coordinates $\mathcal{Y}$ and the pseudo scene coordinates $\mathcal{Y}^*$ in Fig.~\ref{fig:object-placement}. The results indicate that waypoints derived from $\mathcal{Y}$ exhibit higher reprojection errors and less accurate positioning, whereas those from $\mathcal{Y}^*$ achieve more precise positioning, validating the efficacy of the proposed process.

\subsection{Comparison of Virtual Guidance Schemes}
\label{sec:comparison-of-vg}
This section presents a comparison of agent performance under different guidance schemes in the simulation environments, including (a) waypoints, (b) navigation path from $S$ to $D$, and (c) navigation path from $P_t$ to $\omega^*$, as illustrated in Fig.~\ref{fig:virtual_guidance}. In schemes (a) and (b), the routes are planned only once at the beginning of the episode, whereas in scheme (c), the route is updated in real time at each timestep $t$, as detailed in Section~\ref{subsubsec:virtual-guidance-planning}. Table~\ref{tables:impact-analysis-evaluation} presents the agents' performance in the \textit{Hard} environment. The evaluation results indicate that all virtual guidance schemes outperform the baseline. It is worth noting that the agent trained with paths planned from $P_t$ to $w^*$ achieves the best performance. This may be attributed to the virtual guidance consistently directing the agent toward the next waypoint, thereby reducing the chance of losing track. In contrast, agents trained with only waypoints as virtual guidance exhibit the lowest performance among all virtual guidance schemes. This can be explained by the fact that it provides sparse guidance, while the other schemes offer more dense navigational signals.

\section{Conclusion}
\label{sec::conclusion}
In this paper, we introduced virtual guidance as a mid-level representation and validated its effectiveness for sim-to-real policy transfer without fine-tuning. Our experiments demonstrated that agents trained with virtual guidance rendered directly into their observations outperformed those trained with non-visual instructions. We extended this concept by integrating an LLM and an open-vocabulary object detection model to convert language instructions into waypoints. The two-stage process for waypoint placement with pseudo scene coordinates enables accurate positioning of waypoints and the rendering of virtual guidance. The experimental results on a real robot validate that the policies trained in simulation can be transferred to real scenarios.
{
    \small
    \bibliographystyle{ieeenat_fullname}
    \bibliography{main}
}

\end{document}